\documentclass[sigconf]{acmart}

\usepackage{amsfonts}       
\usepackage{amsmath}
\usepackage{multirow}
\usepackage{multicol}
\usepackage{xcolor}

\def\BibTeX{{\rm B\kern-.05em{\sc i\kern-.025em b}\kern-.08emT\kern-.1667em\lower.7ex\hbox{E}\kern-.125emX}}

\renewcommand\footnotetextcopyrightpermission[1]{} 
\begin{document}

%
\title{Learning Patient Engagement in Care Management: Performance vs. Interpretability} 


\author{Subhro Das}
\email{subhro.das@ibm.com}
\orcid{0000-0002xt-7610-2738}
\affiliation{%
  \institution{MIT-IBM Watson AI Lab}
  \streetaddress{}
  \city{IBM Research, Cambridge, MA}
  \country{USA}}

\author{Chandramouli Maduri}
\email{maduric@us.ibm.com}
\affiliation{%
  \institution{IBM T. J. Watson Research Center}
  \streetaddress{}
  \city{Yorktown Heights, NY}
  \country{USA}}

\author{Ching-Hua Chen}
\email{chinghua@us.ibm.com}
\affiliation{%
  \institution{IBM T. J. Watson Research Center}
  \streetaddress{}
  \city{Yorktown Heights, NY}
  \country{USA}}
  
 \author{Pei-Yun S. Hsueh}
\email{phsueh@us.ibm.com}
\affiliation{%
  \institution{IBM T. J. Watson Research Center}
  \streetaddress{}
  \city{Yorktown Heights, NY}
  \country{USA}}



%
\begin{abstract}
The health outcomes of high-need patients can be substantially influenced by the degree of patient engagement in their own care. The role of care managers includes that of enrolling patients into care programs and keeping them sufficiently engaged in the program, so that patients can attain various goals. The attainment of these goals is expected to improve the patients' health outcomes. In this paper, we present a real world data-driven method and the behavioral engagement scoring pipeline for scoring the engagement level of a patient in two regards: (1) Their interest in enrolling into a relevant care program, and (2) their interest and commitment to program goals. We use this score to predict a patient's propensity to respond (i.e., to a call for enrollment into a program, or to an assigned program goal). Using real-world care management data, we show that our scoring method successfully predicts patient engagement. We also show that we are able to provide interpretable insights to care managers, using prototypical patients as a point of reference, without sacrificing prediction performance. 
\end{abstract}

%
%


%
\keywords{Care management, personalization, metric learning, patient-centered care, interpretability, augment intelligence}


%
\maketitle

\section{Introduction}
\label{sec:introduction}

Care management is a patient-centered approach to population health that is ``designed to assist patients and their support systems in managing medical conditions more effectively.'' \cite{chcs2007} The aims of care management decision support (CMDS) may include: (a) identifying populations with modifiable risks, (b) aligning care management services to population needs, and (c) identifying and training personnel to deliver care management services \cite{ahrq2015}. The focus of our paper is on the first of these three aims. To improve the identification of populations with modifiable risks, the Agency for Healthcare Research and Quality (AHRQ) summarized a set of recommendations \cite{ahrq2015}. 
Among the recommendations was that researchers should investigate (a) the benefit of care management services to different patient segments and (b) the parameters that affect modifiable risks. 

With respect to the AHRQ recommendation for achieving the benefits of CMDS, understanding patient segments to drive patient engagement is an inextricable part of the equation for success. By patient engagement, we refer to ``the actions individuals take to obtain the greatest benefit from the health care services available to them.'' \cite{micmrc} 
In addition, it is also essential to develop methods that can identify segment-differentiating parameters that affect modifiable risk factors. These risk factors contributed to a significant portion of global disease burden, especially those with chronic conditions and those who are transitioning from one care setting to another \cite{tuomilehto2001prevention, west1997comprehensive, brown2012six}. The return on investment of care management depends not only on how much the patient stands to gain from a clinical perspective, but also on how likely the patient is to actively engage in care management interventions. 

At the same time, we recognize that, despite its high potential, the adoption of machine learning methods in CMDS scenarios has been slow. This is partly due to the gap between how humans and machines make decisions -- the ``black-box'' nature of high-performing ML methods. To bridge the gap, there is a recent push for more studies in model explainability in AI/ML to help human decision makers understand how the insights were derived and how they can act on the data-driven insights \cite{xai, Caruana2015}. 

As these real-world applications need to work in production environments that often do not come with clearly defined data schema, we will further discuss how to incorporate the developed methods in a Behavioral Engagement Scoring (BES) pipeline (including dynamic feature engineering and API) to enable its applications on care management transaction records in a schema-agnostic fashion.  The pipeline developed is expected to enhance decision support for care managers, by helping them prioritize their efforts based on eventual outcomes/success, but not just clinical health risk. Our methods are informed and validated using real-world care management records for patients who are either transitioning from ``hospital to home'' or are eligible for a chronic disease management program. 

The rest of this paper is organized as follows. We will first discuss how the current study is positioned in the related work in CMDS application and model interpretability. Then, we will introduce the CMDS dataset used in this study, as well as the data-driven methods developed for two CMDS tasks: program enrollment and goal attainment. We will discuss the trade-off between model interpretability and performance observed in the development of machine learning models for identifying explainable engagement behavioral profiles and for personalized engagement scoring from real-world care management data.

\section{Related Work}
\label{sec:RelWork}


In this paper, we focus our investigation on implementing quantitative, data-driven ML methods to identify patient segments who are most likely to engage in care management, and on care manager decision support for explaining for why they may or may not engage. 
While a few quantitative survey-based methods for measuring patient engagement exist \citeN{Graffigna2015,Hibbard2004}, no prior studies have successfully measured patient engagement from real-world data. 

Moreover, although prior studies have attempted to identify risk stratification and disease progression parameters that differentiate clinical risk and longer-term outcomes \cite{Luo2017, Liu2018}, to our knowledge, this is the first study that has proposed to learn engagement strategies from data, based on the understanding of quantifiable difference of patient engagement levels and segment-differentiating parameters that affect modifiable risk factors.  


To bridge the gap between human and machine understanding in the context of CMDS, this study has also explored the potential effect of providing explainable insights on the performance of our models. Recent reviews \cite{Hsueh2017, Lakkaraju2016} have shown that a majority of studies in model interpretability are tied to the optimization of certain model properties that are presumed to be beneficial for improving human understanding of the models. Among them, many studies focused on reducing model complexity. Examples include applying regularization operators to reduce the number of parameters \cite{Feldman2000}, restricting policy search to a subspace of simpler forms \cite{Sridharan2002, Hu2017}, bringing semantically similar items together \cite{Dey2012}, re-training easier-to-understand models to classify on the results obtained by black-box models for model-agnostic explanation \cite{Ribeiro2016}. 

To make AI/ML more ``actionable'' for health decision makers (either health professionals or patients themselves), human-computer interaction researchers have been conducting qualitative studies to identify interpretability-impeding confounders and understand individual differences \cite{Robins1994}. 
With the emergence of deep learning approaches in AI/ML, more studies are now learning patterns that can be represented in explicitly presentable formats (e.g., temporal visualization \cite{Choi2016}, natural language rationalization explanations \cite{Lei2016}), as well as developing interactive tools to untangle models learned in a high-dimensional space \cite{googleexp}.

To further differentiate engagement strategies from CMDS data, in this paper we particularly focus on methods that account for case-based reasoning to improve the interpretability of clustering models, e.g., selecting prototypical cases to represent the learned clusters \cite{Kim2014}. In particular, we incorporate locally supervised metric learning \cite{sun2010localized} and prototypical case-based reasoning in a machine learning model to identify explainable engagement behavioral profiles and to produce personalized engagement scores. 

The main contributions of our paper are as follows: First, we present a quantitative and personalized approach to identifying patients who are more likely to engage in care management, and demonstrate empirically, using real-world data, that our methods provide more accurate engagement behavior predictions compared to a ''one-size-fits-all'' population approach; Second, these insights are made explainable by identifying prototypical patients within a personalized patient segment; we show that, in our case, explainability does not come at the expense of model performance.

\section{Data}
\label{sec:data}

\subsection{Care Management Decision Support}
\label{subsec:DecSuppApp}

For patients with complex care needs, it is important to coordinate across the patients' care givers and providers to account for the differing advice received from clinicians, the varying medications, and the adverse drug events \cite{long2017effective}. In practice, this is often achieved by implementing structured care programs, in which a predetermined set of rules are given to care managers to coordinate with patients and bridge care gaps between hospital and home. Care management history, therefore, captures the important transactions between care managers and patients during the care coordination process, and is an important and growing source of data for behavioral understanding. 

\begin{figure}[t]
  \centering
  \includegraphics[width=\linewidth]{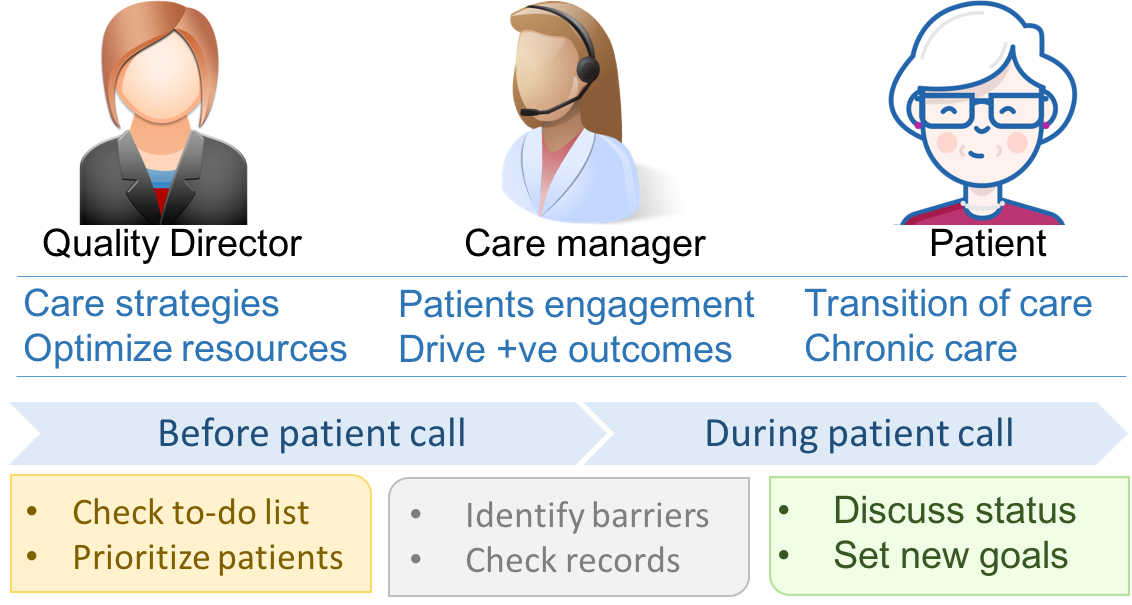}
  \caption{Care Management flow.}
  \Description{Care Management flow.}
  \label{fig:CareMgmnt}
\end{figure}

The CMDS workflow from which this dataset was derived is depicted in Fig. ~\ref{fig:CareMgmnt}. At the center of this figure is the Care Manager (e.g., a licensed nurse, social worker or other certified specialist), who attempts to engage the patient, typically via the telephone, and whose primary objective is to influence modifiable and prioritized risk factors, as identified by the Patient's engagement strategy. The Care Manager receives her assigned pool of patients from the Quality Director, whose primary objective is to align care manager skills with patient needs, and to determine the appropriate care strategies. Finally, the Patient responds to the Care Manager's feedback and coaching and may provide his/her own input on the goals to be set and how to achieve them. The interactions between the care manager and patient are captured in both structure and unstructured format. In the current study, we use only the structured data contained in the care management transaction records. 

\begin{figure}[t]
  \centering
  \includegraphics[width=0.8\linewidth]{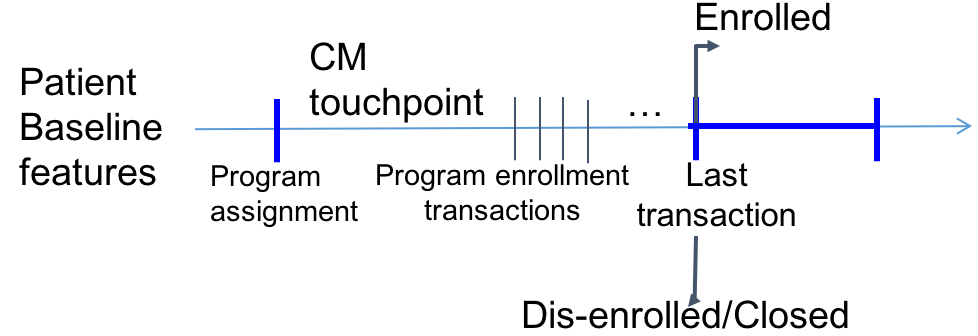} \\
  \caption{Program enrollment timeline}
  \label{fig:ProgTimeline}
  \includegraphics[width=0.8\linewidth]{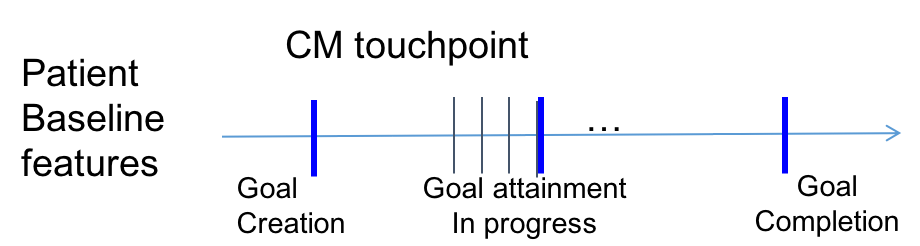}                  
  \caption{Goal attainment timeline.}
  \label{fig:GoalTimeline}
\end{figure}

\subsection{Care Management Records}
\label{subsec:CareMgmtRec}

We apply our method to care program logs of a private, not-for-profit healthcare network, including 4,504 transition of care and 440 chronic care patient interactions over a 22-month period. Those program engagement records were collected between December 2015 and October 2017. For each patient engagement timeline, we extracted 53 features ranging from the basic demographic information (e.g., age, gender), to the patients' care program context (e.g., program experience, whether the patient enrolled in the program, days in the program, number of days until completion of the program) and the interactions between care managers and patients (e.g., the date when the recorded call occurred). 

We then prepared datasets with respect to the two realworld tasks we aim to apply the BES pipeline for decision support: program enrollment (``ENROLL'') and goal attainment (``GOAL'').

\subsection{Program Enrollment}
\label{subsec:dataProgram}

Table~\ref{table:enroll} summarizes the CM records used to generate the ENROLL dataset from all patients  with different enrollment status for each assigned care program. The type of assigned care programs include those transitioning from hospital to home after being discharged from the hospital ("Transition") and those programs involving chronic disease management process ("Involve"). 

\begin{table}[h]
	\centering 
	\begin{tabular}{ | l | c | c | } 
		\hline  
		\rule{0pt}{3ex}  Status & Involve (Chronic Care) & Transition  \\ [1ex] 
		\hline 
		\rule{0pt}{3ex}
		Completed Program & 30.00\% & 67.30\% \\ 
		DidNotEnroll & 4.09\% & 5.71\% \\ 
		Disenrolled & 42.27\% & 25.95\% \\ 
		Enrolled & 23.64\% & 1.04\% \\ [1ex] 
		\hline 
	\end{tabular} 
	\caption{Enrollment status of patients.}
	\label{table:enroll}
\end{table}

\begin{figure}[h]
  \centering
  \includegraphics[width=0.75\linewidth]{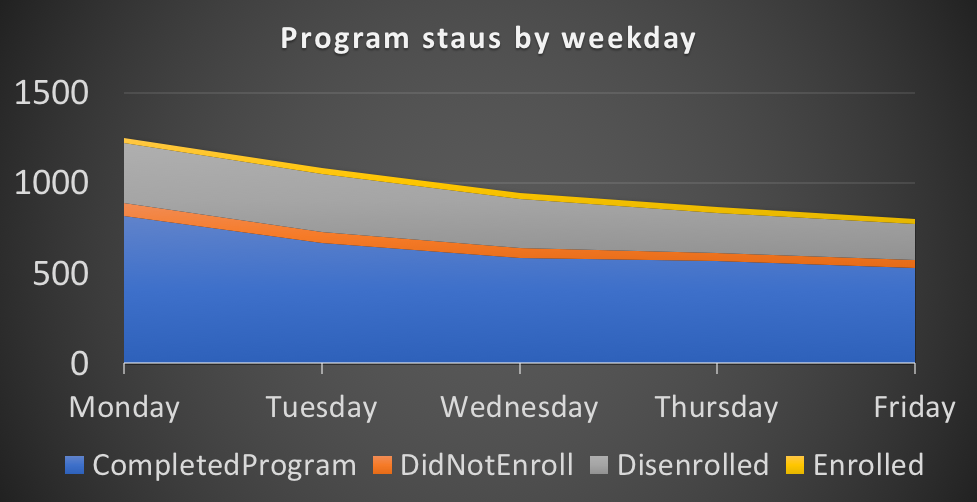}
  \caption{Enrollment distribution over the week.}
  \label{fig:EnrollWeek}
\end{figure}

The structural CM transaction records capture the dates when a care program is assigned, started and ended. These records also contain the indicators of whether the care program is completed with its goals attained, or ended pre-maturely. The assigned program takes 16 days to complete on average and 279 days to complete at the maximum. 
The structural CM records of program enrollment also show that enrolling into the programs on average takes one day and a maximum of 15 days. Fig~\ref{fig:EnrollWeek} summarizes the program enrollment status based on the day of making the recorded call to the target patient. We observed most of the decisions are made in the calls made in the starting of the week.

\subsection{Goal Attainment}
\label{subsec:dataGoal}

\begin{figure}[h]
  \centering
  \includegraphics[width=\linewidth]{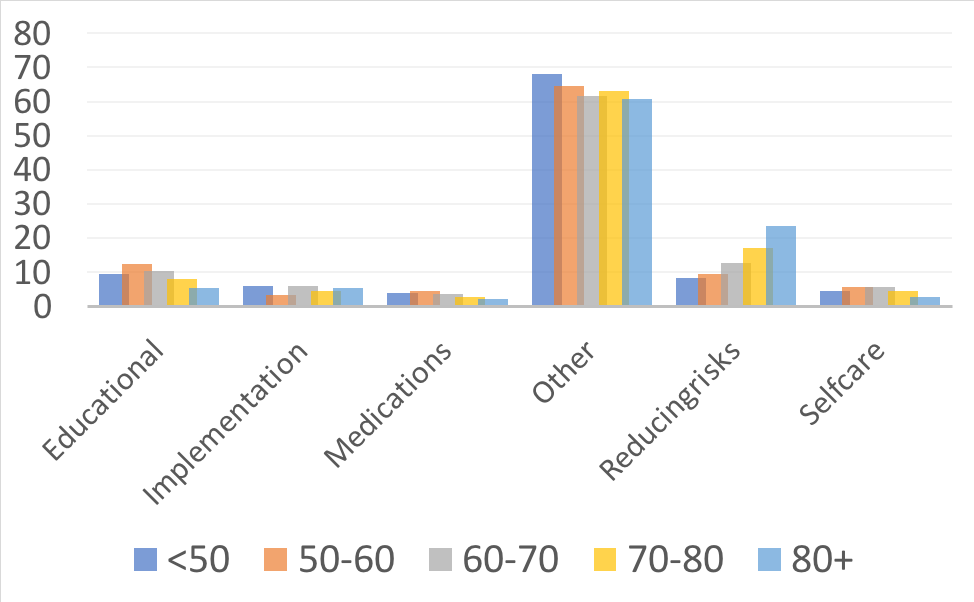}
  \caption{Goal assignment across different focus areas.}
  \label{fig:goalFocusAreas}
\end{figure}

The GOAL dataset is composed of 28 different goals which we classified into six focus areas: Educational  (e.g., demonstrates understanding of post discharge, diabetes education), Implementation (e.g., adequate functional, transportation , support for healthy coping), Medications (e.g., adherence with medication regimen), Reducing Risks (e.g., resolving care gaps), Self Care (e.g., understands benefits of/demonstrates being physically active, healthy diet needs, failure of symptoms management), and Other (e.g., effective care transition and management plans).
Fig~\ref{fig:goalFocusAreas} summarize the goals assigned based on the age category.

Every patient has multiple goals to achieve: 90\% of the patients have fewer than 3 goals, and 65\% patients have fewer than 2 goals. The goal attainment status is indicated by a binary flag (i.e., 1 for meeting the goal and 0 for otherwise).
Table~\ref{table:goalIntervention} summarizes the goal attainment percentage (as indicated by the status shown in the CM records) for each goal focus area, i.e., the number of goals whose status has been shown as 'met' divided by the total number of goals assigned for each key area.

\begin{figure*}[t]
	\centering 
	\begin{tabular}{  l | c  c  c  c  c  c  c  | c | c  } %
		\rule{0pt}{3ex}  Focus Areas & Coaching & Coordination & Education & Referral & Screening & Tracking & Other & Total & Status Met   \\ [1ex] 
		\hline 
		\rule{0pt}{3ex} 
		Educational & 138 & 18 & 277 & 100 & 65 & 0 & 23 & 621 & 81.62\% \\ 
		Implementation & 3 & 192 & 4 & 7 & 0 & 0 & 0 & 206 & 98.54\% \\ 
		Medications & 7 & 96 & 30 & 7 & 90 & 0 &  10 & 240 & 84.91\% \\ 
		Reducing Risks & 0 & 1 & 0 & 545 & 53 & 0 &  1 & 600 & 97.00\%  \\ 
		Self-care & 130 & 0 & 189 & 12 & 116 & 0 & 49 & 496 & 78.69\%  \\ 
		Other  & 29 & 0 & 2561 & 14 & 29 & 12 & 19 & 2644 & 99.19\%  \\ [1ex]
		\hline 
		Total  & 307 & 307 & 3061 & 685 & 353 & 12 & 102 & 4827 & 
	\end{tabular} 
	\caption{Goal attainment records of patients distributed across focus areas \& intervention categories.}
	\label{table:goalIntervention}
\end{figure*}

The interventions of each goal area are grouped into seven categories: Referral (e.g., referral to see a nutritionist for diabetes diet education), Education (e.g., educate patients on the importance of physical activity), Coordination (e.g., follow up with providers on refills), Screening (e.g., assess breathing symptoms), Coaching (e.g., provide a log for side effect recording), and Other (including following up with provider treatment). 

\section{Behavioral Engagement Scoring Pipeline}
\label{sec:methods}

In this paper, we aim to address two key questions: (1) What are the patient segments that lead to the difference of engagement benefits of CM services? (2) What drive differential behavioral responses in CMDS? 
To answer these questions, we develop a Behavioral Engagement Scoring (BES) pipeline to identify the engagement outcome-differentiating factors in care management. The task of engagement scoring serves as a key step in the pipeline to quantify patient engagement tendency for care plan personalization and downstream decision support.

Specifically, the BES pipeline is composed with four components: 1) dynamically extract behavioral features and outcomes based on care management transaction records, 2) apply engagement outcome-driven feature transformation through locally supervised distance metric learning, 3) uncover distinctive patient segments and their behavioral profiles (including prototypical users) based on hierarchical clustering, and 4) learn a BES scorer for each behavioral profile based on a Generalized Linear Model (GLM) to estimate the propensity to respond, e.g., whether a patient is inclined to enroll in a certain program, or complete a goal, given the intervention assigned by his/her care manager.

\subsection{Dynamic Feature Engineering}
\label{subsec:features}

In addition to the main research goal of answering the two key questions, another developmental goal of BES is to enable scalable and flexible engagement scoring over incoming provider data so as to assist in the devising of engagement strategies for real-life CMDS tasks in a production environment. To support this developmental goal, the component of dynamic feature engineering provides a common data model and standard run-time that supports a standards-based analytics environment so as to allow feature generation rules to be written once and used repeatedly with different provider data sources. 

Feature engineering is the process that converts raw data into explanatory factors. These factors are used in the BES pipeline to train engagement scoring models. A multitude of knowledge-based and data-driven approaches are available for feature generation. On one hand, knowledge-based features can be generated from the literature on related topics. On the other hand, data-driven approaches are applied to convert elements of the raw data into features, and then to train models for understanding which features are the most important in gauging patient engagement level. 

In this study, we adopt a hybrid knowledge-augmented, data-driven approach to perform feature engineering. To save time of manual feature generation process and enable better model generalizability across CM data from different providers, we further automate the BES pipeline to generate features in a provider-agnostic fashion, through implementing a back-end logic module with embedded rules that contain knowledge-based rules for converting features from data based on a universal schema coded in configuration files.   

The dynamic feature generation process has further resulted over 700 features. This component has also employed an automatic feature selection procedure based on L1 and L2-based regularization.

\subsection{Engagement Outcome-driven Distance Learning}
\label{subsec:lsml}
The primary motivation to learn an engagement outcome-driven distance metric is to project patient feature-based vectors onto a subspace wherein patients with similar engagement outcomes are closer to each other, whereas those with opposite engagement outcomes are far away from each other. In this transformed vector subspace, we could then further identify cohesive behavioral profiles (using clustering) that drive differential patient responses. 

In this study, we adapt Locally Supervised Metric Learner (LSML) \cite{sun2010localized, sun2012supervised}, which helped estimate the engagement outcome-adjusted distances among patients in the newly transformed vector subspace. The patient features extracted using the protocols mentioned in Subsection~\ref{subsec:features} are represented as $\bf{X} \in \Omega$, where $\bf{\Omega}$ is the vector space, and the class labels $Y \in \{0,1\}$. For the task of program enrollment, the outcome variable is $y=1$, if the status is ''Enrolled'' or ''Completed'', and, $y=0$, if the status is "DidNotEnroll" or "Disenrolled". Similarly, for the task of goal attainment, the outcome variable is $y=1$, if the status is ''Met'', and, $y=0$, otherwise. Considering there are~$n$ program enrollment records and~$d$-dimensional features, then the feature matrix is ${\bf{X}} = \left[ x_1, \cdots, x_i, \cdots, x_n \right] \in \mathbb{R}^{n\times d}$. A similar operational definition is also applicable to goal attainment. 

Here, we consider a generalized Mahalanobis distance, $d_{\Sigma} \left( x_i, x_j \right)$,
\begin{align}
\label{eqn:distance}
d_{\Sigma} \left( x_i, x_j \right)= \left( (x_i - x_j)^T \Sigma (x_i - x_j)  \right)^{\frac{1}{2}}
\end{align}
where, $\Sigma \in \mathbb{R}^{d \times d}$ is a positive semi-definite matrix. We aim to minimize the following distance~$\mathcal{J}$ over the matrix~$\Sigma$:
\begin{align}
\label{eqn:dist_min}
\mathcal{J} = \sum_{i=1}^{n} \left(  \sum_{j:x_j \in \mathcal{N}_i^o} d^2_{\Sigma}\left( x_i, x_j \right)  -  \sum_{k:x_k \in \mathcal{N}_i^e} d^2_{\Sigma}\left( x_i, x_x \right) \right)
\end{align}
where, $\mathcal{N}_i^o$, the homogeneous neighborhood of $x_i$, is the $|\mathcal{N}_i^o|$ nearest data points of $x_i$ with same outcome, and, $\mathcal{N}_i^e$, the heterogeneous neighborhood of $x_i$, is the $|\mathcal{N}_i^e|$ nearest data points of $x_i$ with opposite outcomes. Since~$\Sigma$ is positive semi-definite and symmetric, it can be decomposed as $\Sigma = W^T W$. The~$W^*$ that minimizes~\eqref{eqn:dist_min}, renders the data into the desired space, where records with similar outcomes are compact and those with opposite outcome are distant,
\begin{align}
\label{eqn:W}
W^* = \arg \min_{W : \Sigma = W^T W} \mathcal{J}.
\end{align}
Refer to~\cite{sun2012supervised} for the complete LSML algorithm that derives~$W^*  \in \mathbb{R}^{d \times d}$, the feature transformation matrix. We employed~$W^*$ to obtain the projected feature set,~$\tilde{\bf{X}} \in \tilde{\Omega}$,
\begin{align}
\label{eqn:projX}
\tilde{x_i} = W^*x_i, \quad \forall i.
\end{align}
For the rest of the pipeline, we leverage the outcome-adjusted projection of features ~\eqref{eqn:projX} and the corresponding Mahalanobis distance~\eqref{eqn:distance} between patient-based vectors to learn patient segments and to estimate each patient's propensity to respond to care managers' interventions.

\subsection{Learning Patient Segment-based Behavioral Profiles}
\label{subsec:clustering}
Because we hypothesize that their exists patient segments where within each segment they tend to exhibit certain levels of similarity, we aim to capture that similarity into behavioral profiles and understand engagement-indicative patterns with respect to the profiles they fit in. However, it is not known "a priori" what is the optimal number of patient segments to be clustered into. With that objective and constraint in mind, hierarchical clustering~\cite{friedman2001elements} is employed to identify patient segments on the outcome-adjusted distances in the newly projected vector subspace and learn the key factors that drive the differential engagement outcomes.

Among a variety of linkage methods for hierarchical clustering, we choose Complete Linkage to compute inter-segment similarity~$d_{\text{CL}} \left( G_1, G_2 \right)$ of the furthest pair from segments, say~$G_1$ and~$G_2$, as,
\begin{align}
    d_{\text{CL}} \left( G_1, G_2 \right) &= \max_{x_i \in G_1, x_j \in G_2 } d_{\Sigma^*} \left( x_i, x_j \right), \qquad \Sigma^* = W^{*T} W^* \\
    &= \max_{\tilde{x}_i \in G_1, \tilde{x}_j \in G_2 } d_{\text{euclidean}} \left(\tilde{x}_i, \tilde{x}_j \right).
\end{align}
Experimentation with other linkage methods, e.g., Ward's method, further confirms that the Complete Linkage method uncovers patient segmentation leading to the highest engagement outcome-differentiating power (as measured by ANOVA scores across segments). The remaining challenge is thus to determine the number of segments. As such, an automatic tuning algorithm, Elbow method~\cite{ketchen1996application}, is applied to compute the optimal number of segments. The Elbow method tracks the acceleration of distance growth among segments and thresholds the agglomeration at the point where the acceleration is the highest. Hence, our population is clustered into~$k^*$ segments,~$\mathcal{C} = \{ \mathcal{C}_1, \cdots, \mathcal{C}_{k^*}\}$, each capturing distinctive patterns that drive differential patient responses in engagement. 

\begin{figure*}[h]
  \centering
  \includegraphics[width=\linewidth]{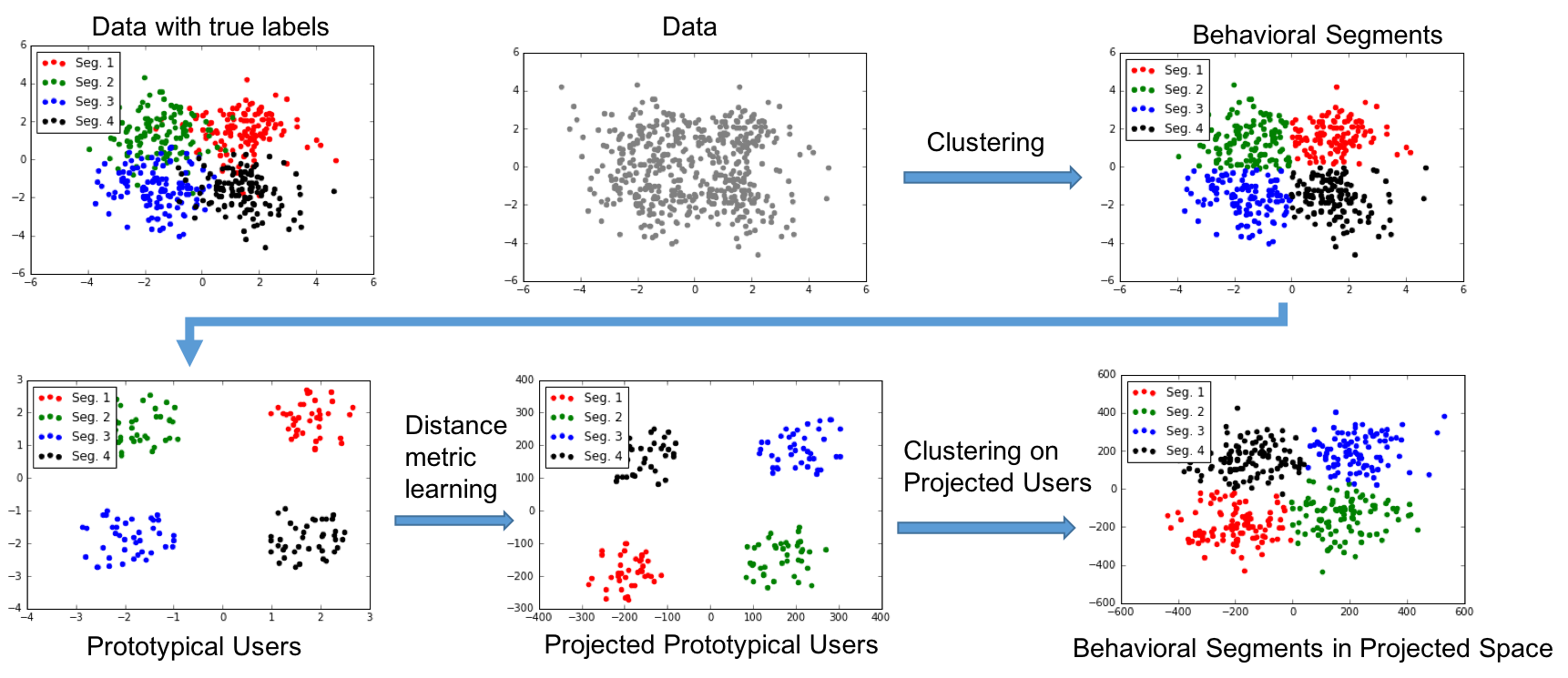}
  \caption{Pipeline evaluation using prototypical patient cases.}
  \label{fig:protoUsers}
\end{figure*}

In addition, as we expect it to be easier to interpret patient need from behavior profiles by examples, we propose to identify prototypical patients in each of the patient segments. The prototypical patient cases in each segment is expected to serve as examples to showcase the distinctive patterns in their behavioral profile. This will help interpret the engagement scores output by the BES pipeline. The prototypical patient cases,~$pU_k$, are defined as the~$p = |pU_k|$ subjects with positive engagement outcome, i.e.,~$y=1$, who are closest to the centroid of each patient segment~$\mathcal{C}_k$ in the engagement outcome-adjusted vector subspace,
\begin{align}
    \label{eqn:protoUsers}
    pU_k & =  \arg \min_{S \; : \; S \subset C_k,  |S| = p, \{ i : y_i = 1 \forall y_i \in S\} } \sum_{\tilde{x}_i \in S} \left( \tilde{x}_i - \mu_{\mathcal{C}_k} \right)^T \left( \tilde{x}_i - \mu_{\mathcal{C}_k} \right), \nonumber \\
    & \qquad\qquad\qquad \forall k = \{ 1, \cdots, k^* \}, 
\end{align}
where,~$\mu_{\mathcal{C}_k}$ is the centroid of the patient segment~$\mathcal{C}_k$. In our analysis, we have chosen~$p=20$. The advantages of having a prototypical case-based component in the pipeline is illustrated in Figure~\ref{fig:protoUsers} using a synthetic set of 2-D data. 

The advantages of using prototypical patient cases include: (a) removing model training noise due to the ambiguous cases near the segment borders, and (b) improving computation efficiency as it takes significantly less run-time to update the models learned on a significantly reduced set of data. 

\subsection{Estimating propensity to respond}
\label{subsec:propensity}
For each of the patient segments projected on the transformed vector subspace, we learn a separate generalized linear model (GLM) to compute engagement scores for each patient in that segment who has been assigned a program to enroll or a goal to attain. The engagement scores are the estimation of each patient's propensity to respond to his/her care manager's engagement calls or interventions. The GLM for each segment~$k$ is represented by
\begin{align}
    \label{eqn:GLM}
    y_i = \beta^k_0 + \beta^k_1 \tilde{x}_{1i} + \cdots + \beta^k_d \tilde{x}_{di}, \quad \forall \tilde{x}_i \in \mathcal{C}_k,
\end{align}
where the feature weights~$\overline{\beta}^k = [\beta^k_0, \cdots, \beta^k_d]$ are computed by minimizing the least squared errors over all the data points from the segment~$k$. Using the optimized feature weights for each segment, the propensity to respond is estimated for each patient based on his/her features. We then use the computed engagement scores to train a Support Vector Machine (SVM) classifier to predict the engagement outcome of each patient. The feature weights of the GLM also provide us more explainable insights specific to each patient segment and to the patients belonging to that segment.

\section{Results \& Discussion}
\label{sec:results}

In this study, we introduce the Behavioral Engagement Scoring pipeline to gauge patient engagement level based on patient segmentation and identify distinctive patterns driving differential responses to engagement. The pipeline is designed to (1) uncover patient segments that lead to the difference of engagement benefits of care management services, (2) identify behavioral profiles that drive differential engagement responses, and (3) enable scalable and flexible engagement scoring in a production environment for real-life care management tasks at each touch point.

The BES pipeline first segment patients based on the patterns exhibited during patient-CM interactions and engagement outcomes. Our hypothesis is that although each feature contains only weak signals to differentiate overall engagement outcomes, when considered collectively in a segment, the combined feature sets can explain rich engagement behaviors for care planning. 

Take the task of program enrollment for example. The BES pipeline first identifies a group of five patient segments from the ENROLL dataset, each of which is found to associate with one behavioral profile of distinctive engagement characteristics. Each segment is then exemplified by incorporating information about the prototypical patient cases (as defined as a subset of top 20 patient cases that are the most representative of the identified segment). The BES pipeline also identifies segment-specific interaction patterns that are specific to program enrollment behaviors for further interpretation and trains GLM models for predicting engagement outcomes. The same is then repeated for training the BES pipeline for the task of goal attainment from the GOAL dataset.


\subsection{Performance evaluation on engagement outcome prediction}
\label{subsec:performance}
\begin{table}[t]
	\centering 
    \begin{tabular}{ | l | l | c | c | l | } %
		\hline  
		\rule{0pt}{3ex}  
		\textbf{Method} & \textbf{Prediction} &  \multicolumn{2}{| c |}{\textbf{Program Enrollment}} & \textbf{Goal}  \\ [1ex] 
		 & \textbf{Performance} &  \multicolumn{2}{| c |}{} & \textbf{Attai}  \\ [1ex] 
		 &  \textbf{Metrics}&  \multicolumn{2}{| c |}{} & \textbf{nment}  \\ [1ex] 
		\hline 
		\rule{0pt}{3ex} 
		&  & \textbf{Involve} & \textbf{Transition} &  \\ 
		 &  & \textbf{(Chronic} &  &  \\ 
		  &  & \textbf{Care)} &  &  \\ 
		\hline 
		\multirow{4}{5em}{\textbf{Population Based (BASELINE)}} & \textbf{Accuracy} & 0.96 & 0.96 & 0.89 \\ 
		\cline{2-5}
		 & \textbf{Precision} & \textcolor{blue}{0.95} & \textcolor{blue}{0.94} & \textcolor{blue}{0.93} \\ 
		 \cline{2-5}
		 & \textbf{Recall} & 0.99 & 1 & 0.95 \\ 
		 \cline{2-5}
		 & \textbf{F1} & 0.97 & 0.97 & 0.94 \\ 
		\hline 
		\multirow{4}{5em}{\textbf{Behavior Profile-Driven}} & \textbf{Accuracy} & 0.93 & 0.74 & 0.79 \\ 
		\cline{2-5}
		& \textbf{Precision} & \textcolor{blue}{0.95} & \textcolor{blue}{0.94} & \textcolor{blue}{0.92} \\ 
		 \cline{2-5}
		 & \textbf{Recall} & 0.96 & 0.65 & 0.83 \\ 
		 \cline{2-5}
		 & \textbf{F1} & 0.95 & 0.76 & 0.87 \\ 
		\hline 
		\multirow{4}{5em}{\textbf{Prototypical User-Driven}} & \textbf{Accuracy} & 0.93 & 0.74 & 0.95 \\ 
		\cline{2-5}
		& \textbf{Precision} & \textcolor{blue}{0.95} & \textcolor{blue}{0.93} & \textcolor{blue}{0.99} \\ 
		 \cline{2-5}
		 & \textbf{Recall} & 0.96 & 0.65 & 0.96 \\ 
		 \cline{2-5}
		 & \textbf{F1} & 0.95 & 0.76 & 0.97 \\ 
		\hline 
	\end{tabular} 
	\caption{Performance comparison with 5-fold cross-validation.}
	\label{table:performance}
	\vskip-20pt
\end{table}

To evaluate the performance of the BES pipeline, the predicted engagement outcome of each patient task is compared with what actually happened as indicated in the care management transaction records. The results across all patient tasks are then aggregated to evaluate the overall performance of models in terms of precision, recall, accuracy and F1-score.    

Two versions of the BES pipeline are evaluated to understand the trade-off between model performance and interpretability. The first "Behavioral Profile-Driven" version trains engagement scoring models for each of the patient segments using all available data belonging to patients in that segment. The second "Prototypical User-Driven" version trains models using only the prototypical patient cases. The performance of the two BES pipelines are also compared with the baseline condition, wherein the "Population-Based" version trains a SVM classification model for engagement outcome prediction using all available data without differentiating patient segments.

Performance evaluation with 5-fold cross validation is shown in Table~\ref{table:performance}. 
Results show that the BES pipeline yields engagement response prediction models of high precision, which implies a high percentage of successful engagements if following the BES recommendations for prioritization. It helps predict patient responses (``whether to engage") for each type of engagement tasks with high precision (>90\%). 

We also explore the potential effect of providing explainable insights on the performance of our models. The precision-based performance metrics of both the Behavioral Profile-Driven and Prototypical User-Driven version are comparatively similar or better than the BASELINE, e.g., training engagement scoring models from the entire population data. The results are encouraging as our proposed BES solution produces more explainable insights based on patient segments and prototypical patient cases, without sacrificing on model performance.
%

\subsection{Drivers of Differential Patient Response} 
\label{subsec:HBIdrivers}
\begin{figure}[t]
  \centering
  \includegraphics[width=\linewidth]{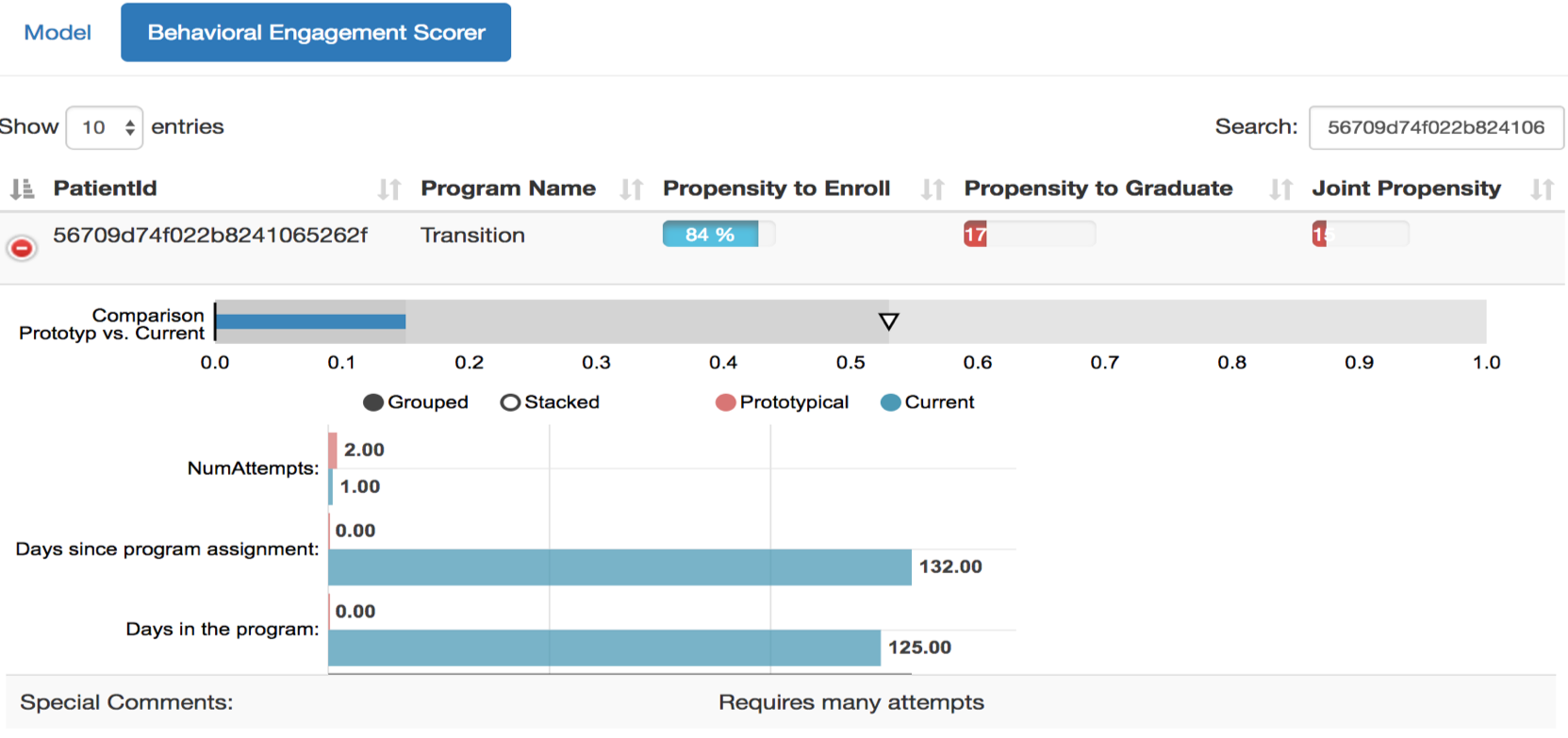}
  \caption{Interpretable engagement insights for Care Managers for shared decision making.}
  \label{fig:BES_demo}
\end{figure}
For each of the two care management tasks, we identify five behavioral profiles for tailoring care management strategies of patient engagement and surface insights for each target patient based on the behavioral profile of his/her closely related patient segment. To achieve this, we analyze feature weights of the model trained for each segment to pinpoint drivers that contribute the most to engagement outcome prediction. This is for generating interpretations for Care Managers to understand the rationale of the predictions offered by the behavioral engagement scorer in the pipeline. 

\begin{figure*}[t]
  \centering
  \includegraphics[width=\linewidth]{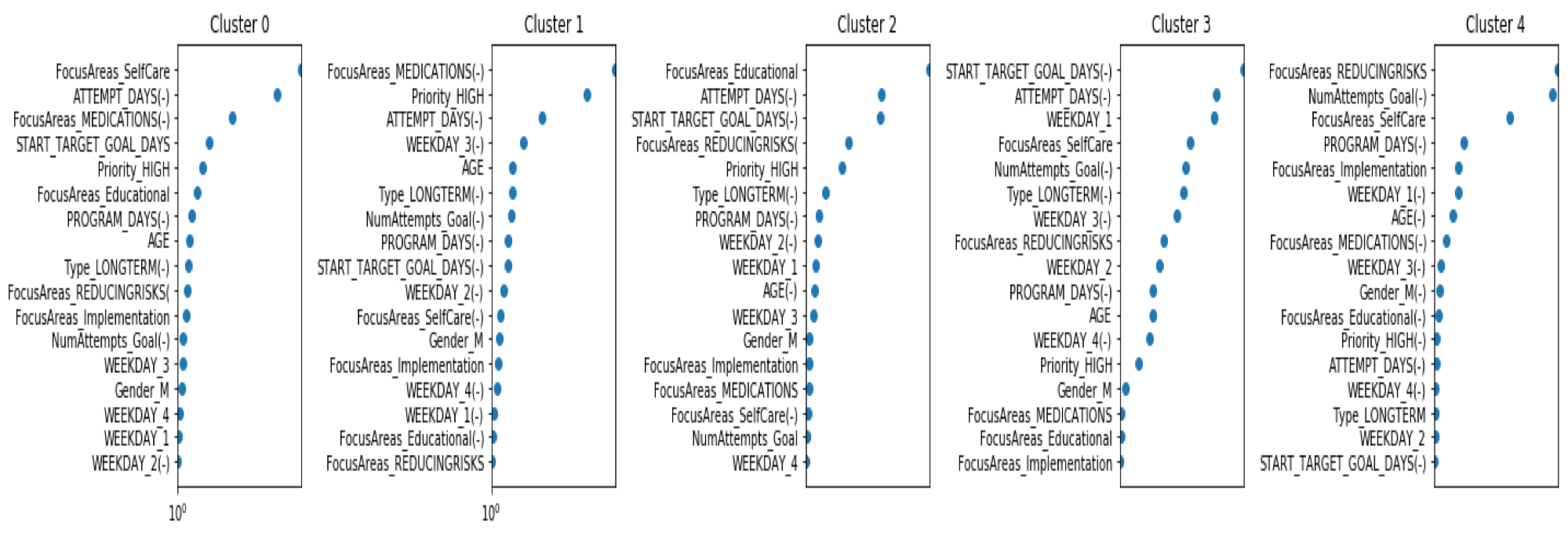}
  \caption{Feature weights that indicate patient response driver across the five different behavioral profiles for goal attainment.}
  \label{fig:featureWts}
\end{figure*}
Fig~\ref{fig:featureWts} demonstrates the variability of differential patient response drivers across the different behavioral profiles for goal attainment. The feature rankings are significantly different when compared the population-level feature rankings with those among patient segments (as indicated by Spearman's coefficient $\rho$;~$p<0.01$). Most of the patient segments exhibit a complex pattern of behavioral response than the population-level ones. 

In some patient segments, we observe strong indicators of care manager influence, e.g., how long the care manager (CM) been trying to help this patient obtain this goal, the number of attempts before goal attainment or before the CM decided to close a goal. In some other patient segments, we observe that certain goals yield more positive engagement responses than the others. For example, in the first and 3rd segment, Self-care and Educational goals are more likely to be attained. The opposite has also been observed in some segments. For example, in the second segment, Medication-related goals are less likely to engage patients in this segment to attain.  Moreover, call context does matter in some segment. For example, for patients in the 4th segment, calling on Tuesday would be more likely to help yield better engagement responses.

It is worthy to note that although we do include age and gender as a part of the features in the analysis. Results show that patient demographics matter less than expected, yielding only neutral contribution to the modeling of engagement response prediction.

\subsection{Interface and API for Shared Decision Making based on Engagement Insights } 
\label{subsec:API}

The BES pipeline also includes a web-based user interface that provides access to explainable insights about the gauged engagement level and the predicted response.  In addition, the pipeline derives best practice insights using the example illustrated by the prototypical patient cases in each segment.  Care managers can use the interface to learn about the target patients and their related patient segments for prioritization and best practice learning. 

Fig~\ref{fig:BES_demo} shows the interactive tooling based on a demo API customized for care managers to make their decisions before the call regarding which patients to call first for program enrollment and for goal attainment, and what are the reasons that they might or might not be engaged.

\section{Conclusion \& Future Work}
\label{sec:discussion}

The main contributions of our paper are as follows: First, we present a quantitative and personalized approach to identifying patients who are more likely to engage in care management, and demonstrate empirically, using real-world data, that our methods provide more accurate engagement behavior predictions compared to a ``one-size-fits-all'' population-based approach; Second, these insights are made explainable by identifying prototypical patient cases within a personalized patient segment. Performance evaluation results show that, in our case, explainability does not come at the expense of model performance.

Analyzing observational transaction data of care management interaction logs regarding to clinical factors only overlooks an important part of equation leading to better outcomes. It is expected that segment-level incidence rates might result in biased estimates of the effect of interventions. Applying the BES pipeline, which properly adjusts for individual and patient segment information, enables a more accurate estimate of engagement effect and support care management decision-making and in a shared decision-making scenario. The quantification of heterogeneous engagement effects in patient segments goes beyond the existing care quality metrics to add another perspective of behavioral understanding to providers using care management programs.

As for the issue of bridging the gap between human and machine decision making, simply optimizing model properties is not sufficient to warrant actions from health decision makers \cite{Chen2017}. More research is needed to identify additional evaluation metrics that can serve as a proxy measure of the performance in real-life user tasks \cite{Karkar2015}. This line of research can help evaluate how to make sense of the models and analytical results in order to support decision makers' actions, as well as how to validate the derived insights directly to automate decisions. Doing this is a truly interdisciplinary work and expected to enhance future milestones to develop tools for creating deployment and feedback process and aligning with the need of generating real-world evidence on best practice.

%


%
\bibliographystyle{ACM-Reference-Format}
\bibliography{references}




\end{document}